\documentclass[sigconf]{acmart}

\copyrightyear{2023} 
\acmYear{2023} 
\setcopyright{rightsretained} 
\acmConference[CIKM '23]{Proceedings of the 32nd ACM International Conference on Information and Knowledge Management}{October 21--25, 2023}{Birmingham, United Kingdom}
\acmBooktitle{Proceedings of the 32nd ACM International Conference on Information and Knowledge Management (CIKM '23), October 21--25, 2023, Birmingham, United Kingdom}\acmDOI{10.1145/3583780.3614819}
\acmISBN{979-8-4007-0124-5/23/10}

\usepackage{xspace}
\usepackage{enumitem}
\usepackage{xcolor}

\newcommand{\model}[0]{\textsc{ConceptEvo}\xspace}
\newcommand{\basemodel}[0]{base model\xspace}

\definecolor{red}{RGB}{198,50,42}
\definecolor{agreen}{RGB}{74, 198, 148}
\definecolor{purple}{RGB}{158, 62, 177}
\definecolor{aqua}{RGB}{87, 180, 181}
\definecolor{orange}{RGB}{255,143,40}
\definecolor{amber}{rgb}{1.0, 0.75, 0.0}
\definecolor{awesome}{rgb}{1.0, 0.13, 0.32}
\definecolor{bronze}{rgb}{0.8, 0.5, 0.2}
\definecolor{indigo}{rgb}{0.0, 0.25, 0.42}
\definecolor{heliotrope}{rgb}{0.87, 0.45, 1.0}
\definecolor{forestgreen}{rgb}{0.13, 0.55, 0.13}
\definecolor{ginger}{rgb}{0.69, 0.4, 0.0}
\definecolor{jade}{rgb}{0.0, 0.66, 0.42}
\definecolor{mediumslateblue}{rgb}{0.48, 0.41, 0.93}
\definecolor{mint}{rgb}{0.24, 0.71, 0.54}
\definecolor{mulberry}{rgb}{0.77, 0.29, 0.55}
\definecolor{linkColor}{RGB}{6,125,233}

\newcommand{\add}[1]{\textcolor{black}{#1}}

\begin{document}

\title{Concept Evolution in Deep Learning Training:\\ A Unified Interpretation Framework and Discoveries}

\author{Haekyu Park}
\affiliation{
    \institution{Georgia Tech}
    \country{ }
}
\email{haekyu@gatech.edu}

\author{Seongmin Lee}
\affiliation{
    \institution{Georgia Tech}
    \country{ }
}
\email{seongmin@gatech.edu}

\author{Benjamin Hoover}
\affiliation{
    \institution{Georgia Tech}
    \country{ }
}
\email{bhoov@gatech.edu}

\author{Austin P. Wright}
\affiliation{
    \institution{Georgia Tech}
    \country{ }
}
\email{apwright@gatech.edu}

\author{Omar Shaikh}
\affiliation{
    \institution{Georgia Tech}
    \country{ }
}
\email{oshaikh@gatech.edu}

\author{Rahul Duggal}
\affiliation{
    \institution{Georgia Tech}
    \country{ }
}
\email{rahulduggal@gatech.edu}

\author{Nilaksh Das}
\affiliation{
    \institution{Georgia Tech}
    \country{ }
}
\email{nilakshdas@gatech.edu}

\author{Kevin Li}
\affiliation{
    \institution{Georgia Tech}
    \country{ }
}
\email{kevin.li@gatech.edu}

\author{Judy Hoffman}
\affiliation{
    \institution{Georgia Tech}
    \country{ }
}
\email{judy@gatech.edu}

\author{Duen Horng Chau}
\affiliation{
    \institution{Georgia Tech}
    \country{ }
}
\email{polo@gatech.edu}

\renewcommand{\shortauthors}{Park et al.}






\begin{abstract}
We present \model, a unified interpretation framework for deep neural networks (DNNs) that reveals the inception and evolution of learned concepts during training.
Our work addresses a critical gap in DNN interpretation research, as existing methods primarily focus on post-training interpretation.
\model introduces two novel technical contributions: 
(1) an algorithm that generates a unified semantic space, enabling side-by-side comparison of different models during training,
and 
(2) an algorithm that discovers and quantifies important concept evolutions for class predictions.
Through a large-scale human evaluation and quantitative experiments, we demonstrate that \model successfully identifies concept evolutions across different models, which are not only comprehensible to humans but also crucial for class predictions.
\model is applicable to both modern DNN architectures, such as ConvNeXt, and classic DNNs, such as VGGs and InceptionV3. 
\end{abstract}

\maketitle

\section{Introduction}

\begin{figure}[t!]
    \centering
    \includegraphics[width=1\columnwidth]{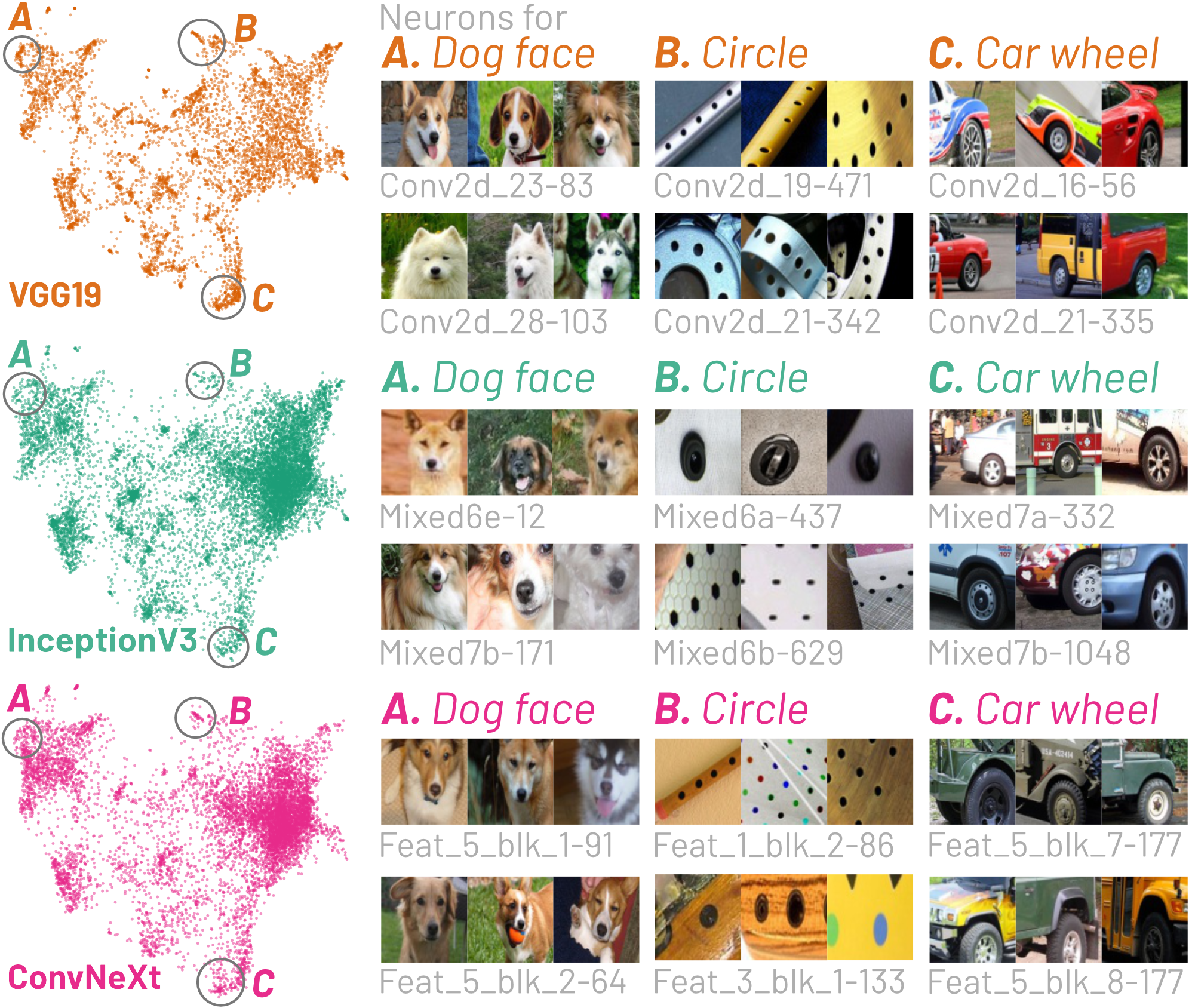}
    \vspace{-1.5em}
    \caption{
        \model creates a unified semantic space 
        that enables side-by-side comparison of different models during training (top: VGG19; middle: InceptionV3; bottom: ConvNeXt).
        \model embeds and aligns neurons (dots) that detect similar concepts (e.g., dog face, circle, car wheel) to similar locations. 
    }
    \vspace{-1em}
    \label{fig:neuron-alignment}
\end{figure}

Interpreting how Deep Neural Networks (DNNs) arrive at their decisions has become crucial for instilling trust in the models~\cite{ribeiro2016should}, debugging them~\cite{koh2017understanding}, and guarding against potential harms such as embedded bias or adversarial attacks~\cite{das2020bluff,papernot2018deep,zhang2018examining}.
As a fundamental type of DNN, convolutional neural networks have garnered significant interest in understanding their internal mechanism.
Saliency-based interpretation methods, for example, aim to identify important image regions for predictions~\cite{selvaraju2017grad,simonyan2013deep}. 
Concept-based interpretation methods identify \textbf{\textit{concepts}} detected by DNNs, such as ``\textit{dog face}'' concepts shown in Fig \ref{fig:neuron-alignment}, and their role in forming higher-level concepts and predictions~\cite{park2021neurocartography,olah2020zoom,ghorbani2019towards,kim2018interpretability,bau2017network}. 
These methods connect a concept with sets of images or image patches that explain the concept, using shared visual characteristics among the images to enhance human understanding of the concept \cite{chen2019looks, olah2017feature, ghorbani2019towards}.
Neuron-level concept interpretation methods focus on concepts that elicit strong activation in that neuron \cite{olah2017feature, chen2019looks, park2021neurocartography}.

However, existing interpretation approaches mostly focus on post-training analysis~\cite{laugel2019dangers,guidotti2018survey}, 
providing limited insights into the evolution of models during training.
Crucially, understanding the progression of concepts detected by individual neurons, which we refer to as the neuron's \textbf{concept evolution}, and its association with model deficiencies like poor generalizability \cite{li2018visualizing, zhang2021understanding, keskar2016large} or convergence failures \cite{reddi2019convergence, arora2018convergence} remains lacking.
Relying solely on post-training interpretation poses challenges for real-time discovery and diagnosis during training, potentially wasting time and resources \cite{elsken2019neural, safarik2018genetic}, if the training ultimately fails to achieve desired outcomes.
Interpreting the DNN training process also enhances effective monitoring \cite{zhong2017evolutionary,liu2017analyzing,abadi2016tensorflow, zhou2022neuromapper}.

\begin{figure}[t]
    \centering
    \includegraphics[width=1\columnwidth]{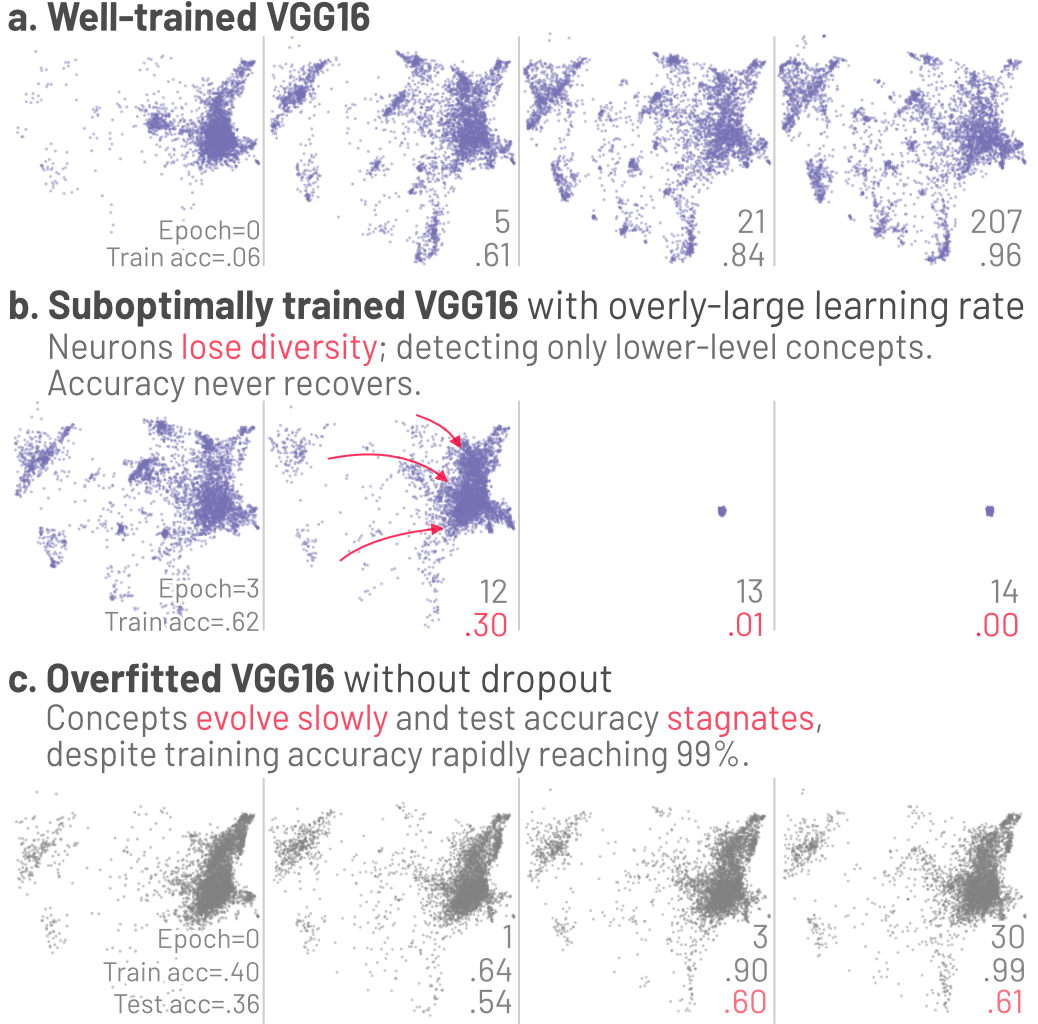}
    \vspace{-2em}
    \caption{
        \model identifies potential training issues.
        (a) A well-trained VGG16 shows gradual concept formations and refinements.
        (b) A VGG16 suboptimally trained with a large learning rate, rapidly losing the ability to detect most concepts.
        (c) An overfitted VGG16 without dropout layers, showing slow concept evolutions despite rapid training accuracy increases. 
        We abbreviate ``top-5 training/test accuracies'' as ``train/test acc.''
    }
    \vspace{-2em}
    \label{fig:discovery}
\end{figure}

To fill these gaps, our work contributes as follows:

\begin{enumerate}[topsep=2pt, itemsep=0mm, parsep=2pt, leftmargin=10pt,label*=\textbf{\arabic*.}]

    \item \textbf{\model, a unified interpretation framework that reveals the inception and evolution of concepts during DNN training} (Sec~\ref{sec:method}), with two novel technical contributions\footnote{\model has been made open source: \url{https://github.com/poloclub/ConceptEvo}.}:
        \begin{itemize}[leftmargin=3mm, topsep=0mm, parsep=0.5mm, itemsep=0mm]
            \item 
                An algorithm that generates a unified semantic space that enables side-by-side comparison of different models during training (Fig \ref{fig:neuron-alignment}, \ref{fig:discovery}).
                \model is applicable to both modern ConvNeXt and classic DNNs like VGGs and InceptionV3.
            \item 
                An algorithm that discovers and quantifies important concept evolutions for class predictions (Fig \ref{fig:important-evo}). 
        \end{itemize}

    \item \textbf{Extensive evaluation} (Sec \ref{sec:experiment}).
    A large-scale human experiments with 260 participants and quantitative experiments demonstrate that \model identifies concept evolutions that are not only meaningful to humans but also important for class predictions.

    \item \textbf{Discoveries on model evolution} (Sec \ref{sec:discovery}). 
    We highlight how \model aids in uncovering potential issues during model training and provides insights into their causes, such as:
    (1) severely harmed concept diversity caused by
    incompatible hyperparameters (e.g., overly high learning rate) as shown in Fig~\ref{fig:discovery}b; and
    (2) slowly evolving concepts despite rapid increases in training accuracy in overfitted model as shown in Fig~\ref{fig:discovery}c.

\end{enumerate}

\section{Related Work}

\textbf{Interpreting DNNs After Training.}
Interpreting fully-trained DNNs revolves around describing crucial features of models' behavior.
For example, saliency-based methods identify image pixels that are important for predictions \cite{simonyan2013deep, simonyan2015very, gan2015devnet, selvaraju2017grad}.
However, these methods face a challenge as important image pixels may not align with high-level concepts that are easily understandable to humans \cite{kim2018interpretability, gulshad2020explaining}.
To address this, recent studies have focused on explaining high-level, human-understandable concepts learned within DNNs and their relevance to the models' prediction \cite{hernandez2022natural, ghorbani2020neuron, YehKALPR20, GoyalSK19, zhou2018interpreting, kim2018interpretability, NguyenYC16}.
For example, feature visualization techniques \cite{zeiler2014visualizing, yosinski2015understanding} generate synthetic images that strongly activate specific neurons, visualizing detected concepts.
ACE \cite{ghorbani2019towards} discovers important image segmentations, presenting learned concepts that are important for predictions.
Net2Vec~\cite{FongV18} encodes individual neurons' concepts into vectors by using predefined concept images. 
MILAN~\cite{hernandez2022natural} explains learned concepts through short natural language descriptions.
NeuroCartography~\cite{park2021neurocartography} visualizes concepts detected by neurons through encoding the conceptual neighborhood of neurons.

\textbf{Interpreting DNNs During Training.}
Several existing studies that aim to interpret DNNs \textit{during training} focus on the evolution of data representations within the models across epochs and how this evolution influences their downstream performance \cite{puhringer2020instanceflow, smilkov2017direct, chung2016revacnn}.
DeepEyes \cite{pezzotti2017deepeyes} examines the evolution of individual neurons' activation for different classes during training.
DGMTracker~\cite{liu2017analyzing} analyzes changes in weights, activations, and gradients over time.
Other approaches track the 2D projected evolution of neurons towards or away from specific labels~\cite{rauber2016visualizing, li2020visualizing}, although this limits our understanding of learned concepts to the available labels only. 
DeepView~\cite{zhong2017evolutionary} introduces metrics to estimate whether neurons are evolving sufficient diversity for classification.
\model distinguishes itself from the existing approaches by enabling the comparison of concepts learned by neurons from \textbf{any layer within a model} and even by neurons from \textbf{different models}.

\section{Method}
\label{sec:method}

\subsection{Desiderata of Interpreting Concept Evolution}
\label{sec:desiderata}

\begin{enumerate}[label=\textbf{D\arabic*},itemsep=0mm, topsep=1mm,parsep=1mm, leftmargin=6mm]

    \item \textbf{General interpretation of concept evolution across different models.} \label{d:comparable}
        Comparing the training of different models is essential for determining which model is trained better or which training strategy is more effective  \cite{li2018visualizing, raghu2017svcca}.
        Thus, we aim to develop a general method that enables side-by-side comparison and interpretation of concept evolution across different models.
        (Sec~\ref{sec:comparable})
        
    \item \textbf{Revealing and quantifying important evolution of concepts.} \label{d:important}
        We aim to identify internal changes that significantly impact the prediction of a specific class, as understanding the most influential components can lead to effective model improvements \cite{ghorbani2020neuron}.
        For example, we seek to determine the importance of a neuron's concept evolution, such as the transition from ``brown color'' to ``brown furry leg'' in the prediction of a ``brown bear'' class.
        We aim to automatically discover these important changes in concepts for class predictions. (Sec~\ref{sec:important})
    
    \item \textbf{Discoveries.} 
        \label{d:actionable}
        Can the interpretation of how a model evolves help identify training problems and provide insights for addressing them, advancing prior work that focuses on interpreting and fixing models post-training \cite{ghorbani2020neuron}?
        For example, can we help determine if a model's training is on the right track and if interventions are necessary to improve accuracy?
        (Sec~\ref{sec:discovery})
        \vspace{-0.5em}
\end{enumerate}

\subsection{General Interpretation of Concept Evolution}
\label{sec:comparable}

We desire an interpretation of model evolution that is comparable across different models.
However, direct comparison between concepts in different models at different training stages is challenging.
Different models are independently trained; thus, the learned concepts are not aligned by default.
Even for the same model, activation patterns can change considerably over training epochs.

To address this challenge, we propose a two-step method.
In step~1, we create a base semantic space that captures the concepts identified by a \textit{\basemodel} at a specific training epoch. 
\add{This semantic space serves as a fundamental reference for concept representation.}
In step 2, we project the concepts from other models spanning all epochs onto the base semantic space, resulting in a \textbf{unified semantic space} where similar concepts across different models and epochs are mapped to similar locations.

We choose an optimally, fully trained model as our \basemodel to ensure broad concept coverage.
For example, we used a fully trained VGG19~\cite{simonyan2015very} as the \basemodel for Fig \ref{fig:neuron-alignment} and \ref{fig:discovery}.

\textbf{Step 1: Creating the base semantic space.}
To create the \textit{base semantic space}, we use neurons as a unit to identify and represent concepts, inspired by 
studies that demonstrate neurons' selective activation for specific concepts \cite{ghorbani2020neuron, olah2017feature, yosinski2015understanding}.
By using neurons, we can pinpoint areas of interest in models, enabling focused troubleshooting, particularly in identifying abnormal training patterns within specific groups of neurons.
Building on prior work \cite{park2021neurocartography}, we embed neurons that strongly respond to common inputs in similar locations.
As neuron-concept relationships may not always be one-to-one~\cite{olah2020zoom, FongV18}, we aim to generalize to many-to-many relationships. 
For example, polysemantic neurons responsive to multiple concepts are embedded between those concepts.

\textbf{Step 1.1: Finding stimuli.} 
\model creates stimuli for each neuron by collecting a set of $k$ images that result in the highest maximum in the neuron's activation map.
For neurons associated with a single concept, their stimuli will be more alike, while for polysemantic neurons, their stimuli may consist of multiple concepts.

\textbf{Step 1.2: Sampling frequently co-activated neuron pairs.} 
\model creates a multiset $D$, which consists of sampled pairs of strongly co-activated neurons from the \basemodel $M_b$ at epoch $t_b$. 
First, for each image $\mathbf{x}$, it creates a list of neurons that are strongly co-activated by $\mathbf{x}$, by collecting neurons with $\mathbf{x}$ in their stimuli.
Next, it randomly shuffles each list of co-activated neurons and samples neuron pairs using a sliding window of length two over the shuffled neurons.
The sampled neuron pairs are added to $D$.
This sampling process is repeated $E$ times to obtain diverse neuron pairs. 
Note that a specific neuron pair can appear multiple times in $D$, with their frequency of appearance increasing as more images are shared by their stimuli. 
This leads to a closer embedding of more frequently co-activated neurons in the unified semantic space.

\textbf{Step 1.3: Learning neuron embedding.} 
The objective function, defined by Eq~\eqref{eq:neuronemb}, represents a negative log likelihood to learn neuron embeddings;
\add{intuitively, 
(1) co-activated neuron pairs with a larger inner product (and spatially closer embeddings) are more likely to indicate similar concepts, while 
(2) randomly paired neurons with a lower inner product) and spatially farther embeddings) are less likely to be conceptually similar.
The randomly paired neurons serve as negative examples, enabling high-quality vector representations of concepts, similar to the negative sampling approach used in Word2Vec algorithm \cite{mikolov2013efficient, mikolov2013distributed}.
This neuron embedding approach allows for the representation of many-to-many relationships between neurons and concepts.
For example, a polysemantic neuron, which is co-activated by multiple distinct groups of neurons representing different concepts, is attracted towards these groups, resulting in its spatial location between them.}
In the objective function, $\mathbf{v}^{t}_{n,M}$ is an embedding of neuron $n$ in model $M$ at epoch $t$. 
$r$ is a randomly selected neuron.
$R$ is the number of randomly sampled neurons for each co-activated neuron pair in $D$.
$\sigma(\cdot)$ is the sigmoid function (i.e., $\sigma(x) = 1 / (1 + e^{-x})$). 
\begin{equation}
    \label{eq:neuronemb}
    \begin{split}
    & J_1 =
      -\displaystyle \sum_{(n, m) \in D} \bigg(
        \log \big(\sigma(\mathbf{v}^{t_b}_{n,M_b} \cdot \mathbf{v}^{t_b}_{m,M_b}) \big) \; +
            \\ 
            & \sum_{r=1}^{R}
                \log \big (1-\sigma(\mathbf{v}^{t_b}_{n,M_b} \cdot \mathbf{v}^{t_b}_{r,M_b}) \big ) 
            + \sum_{r=1}^{R} 
                \log \big (1-\sigma(\mathbf{v}^{t_b}_{m,M_b} \cdot \mathbf{v}^{t_b}_{r, M_b}) \big )
    \bigg)
    \end{split}
\end{equation}
\vspace{-1em}

We randomly initialize the neuron embeddings and learn the embeddings by gradient descent.
Eq~\eqref{eq:derivative-neuronemb-n} and \eqref{eq:derivative-neuronemb-m} present the derivative to update the neuron embeddings.
\vspace{-0.5em}
\begin{equation}
    \begin{split}
    \label{eq:derivative-neuronemb-n}
    \frac{\partial J_1}{\partial \mathbf{v}^{t_b}_{n,M_b}} 
         =
        (1 - \sigma(\mathbf{v}^{t_b}_{n,M_b} \cdot \mathbf{v}^{t_b}_{m,M_b})) 
            \mathbf{v}^{t_b}_{m,M_b} 
        - \displaystyle \sum_{r=1}^{R} 
        \sigma(\mathbf{v}^{t_b}_{n,M_b} \cdot \mathbf{v}^{t_b}_{r,M_b}) \;
            \mathbf{v}^{t_b}_{r,M_b}
    \end{split}
\end{equation}
\vspace{-1em}
\begin{equation}
    \begin{split}
    \label{eq:derivative-neuronemb-m}
    \frac{\partial J_1}{\partial \mathbf{v}^{t_b}_{m,M_b}} 
        =
        (1 - \sigma(\mathbf{v}^{t_b}_{n,M_b} \cdot \mathbf{v}^{t_b}_{m,M_b})) \; 
            \mathbf{v}^{t_b}_{n,M_b}
        - \displaystyle \sum_{r=1}^{R} 
        \sigma(\mathbf{v}^{t_b}_{m,M_b} \cdot \mathbf{v}^{t_b}_{r,M_b}) \;
            \mathbf{v}^{t_b}_{r,M_b}
    \end{split}
\end{equation}

\textbf{Step 2: Unifying the semantic space of different models at different epochs.}

\textbf{Step 2.1: Image embedding}. 
\add{Different models, with varied architectures and neurons, can share the commonality of being trained on the same dataset. 
Leveraging this, we consider that if two neurons from different models are strongly activated by the same inputs, they likely detect the same concept.
To represent neurons' concepts across models, we use image embeddings as a bridge: 
we compute image embeddings that approximate the original neuron embeddings in the \basemodel, and these image embeddings are then used to approximate the neuron embeddings in other models.}

A neuron's embedding typically represents a more detailed concept (e.g., car wheel as shown in Fig \ref{fig:neuron-alignment}) extracted from the entire images (e.g., car images) that include various concepts (e.g., car wheels, loads, and more).
Thus, we consider that collective embeddings of neurons can approximate the image embedding.
Similarly, we assume that a neuron's embedding can be formed by collectively considering the embeddings of images to which the neuron strongly responds,
In particular, we aim to encode a common concept (e.g., car wheel) across the stimuli (e.g, car images) into the neuron's embedding.
To approximate a neuron's embedding, we consider linearly combining the embeddings of the stimuli of the neuron, reinforcing the common concepts (e.g., car wheel) by summing the shared features encoded in the image embeddings.
Unrelated concepts (e.g., backgrounds or different colors of cars) which may occur randomly and vary in presence (or absence) across stimuli can be disregarded by summing and zeroing out such unrelated concepts' (positive and negative) contributions.
To aggregate the embeddings, we adopt the standard practice of averaging across the important images as in previous seminar work~\cite{ghorbani2020neuron, ghorbani2019towards, kim2018interpretability}.
Eq~\eqref{eq:neuron-embedding-approximation} presents the neuron embedding approximation, where
$X^{t_b}_{n, M_b}$ is the set of stimuli of neuron $n$ in the \basemodel $M_b$ at epoch $t_b$.

\begin{equation}
    \mathbf{v}'^{t_b}_{n,M_b}=
        \frac{1}{|X_{n,M_b}^{t_b}|}
        \sum_{\mathbf{x}\in X_{n,M_b}^{t_b}} \mathbf{v}_\mathbf{x}
    \label{eq:neuron-embedding-approximation}
\end{equation}

Eq~\eqref{eq:objective-img} presents the objective function to minimize the difference between the original and the approximated embedding of neurons in the \basemodel, where $N_{M_b}$ is a set of all neurons in the \basemodel.
We randomly initialize the image embeddings and learn them by gradient descent. 
Eq~\eqref{eq:derivative-img-emb} shows the derivative used to update an image's embedding, where $N_{M_b,\mathbf{x}}$ is the set of neurons in $M_b$ whose stimuli includes an input~$\mathbf{x}$.

\begin{equation}
    \label{eq:objective-img}
    J_2 = 
        \frac{1}{2}
        \displaystyle \sum_{n \in N_{M_b}} 
        \bigg|\bigg| 
            \mathbf{v}'^{t_b}_{n,M_b} - \mathbf{v}^{t_b}_{n,M_b}
        \bigg|\bigg|_2^2
\end{equation}

\begin{equation}
    \label{eq:derivative-img-emb}
    \frac{\partial J_2}{\partial \mathbf{v}_\mathbf{x}} = 
    \displaystyle
    \sum_{n \in N_{M_b,\mathbf{x}}}
        \frac{1}{|X^{t_b}_{n, M_b}|}
        \bigg(
            \mathbf{v}'^{t_b}_{n,M_b}
            - \mathbf{v}^{t_b}_{n,M_b}
        \bigg)
\end{equation}

The image embedding approach may have a limitation as it can only represent images from the top-$k$ stimuli of neurons in the \basemodel.
Consequently, if none of the images in a neuron's stimuli are not covered by the \basemodel, the neuron itself remains unrepresented.
With a large number of images, the top-$k$ sets of stimuli for two models may have a low chance of overlapping. 
To address this issue, we use a randomly sampled images (10\% sampled) instead of using all of them to increase the chance of overlapping.
Additionally, we indirectly represent images that are not covered by the \basemodel's stimuli by adopting a similar approach as in Step 1;
instead of representing neurons based on their co-activation by common images,
we represent images based on how they make common neurons co-activated.
For each image $\mathbf{x}$, \model identifies the $k$ most activated neurons by $\mathbf{x}$, denoted as $N^{t_b}_{ M_b, \mathbf{x}}$.
Images $\mathbf{x}_1$ and $\mathbf{x}_2$ are paired if there are common neurons in $N^{t_b}_{M_b, \mathbf{x}_1}$ and  $N^{t_b}_{M_b, \mathbf{x}_2}$.
The paired images are added to the multiset of image pairs denoted as $S$. 
Image pairs in $S$ may appear more than once (i.e., $S$ is a multiset), indicating that those images can stimulate more common neurons, leading to a closer embedding.
The image embeddings are learned in a similar manner to the neuron embedding approach, with the embeddings for images that are already represented by the \basemodel being fixed.

\textbf{Step 2.2: Approximating embedding of neurons in other models at different epochs}. 
After embedding images in Step 2.1, \model approximates neuron embeddings of other models at other epochs by averaging the embedding of images in each neuron's stimuli that are covered by the \basemodel.
If none of the images in a neuron's stimuli are covered by the \basemodel, it averages the indirectly derived image embeddings.
Step 2.2 is the only necessary (sub)step when projecting concepts in a new model onto the unified semantic space. There is no need to repeat Step 1 and Step 2.1.

To visualize the neuron embeddings, we use UMAP, a non-linear dimensionality reduction method that preserves both the global data structures and local neighbor relations \cite{mcinnes2018umap}.
To assist in understanding the concepts that neurons strongly respond to, we compute example patches which are cropped images that maximally activate the neuron (e.g., example patches of neurons for the \textit{``dog face''} concept in Fig~\ref{fig:neuron-alignment}) \cite{olah2017feature}.

\subsection{Concept Evolutions Important for a Class}
\label{sec:important}

Our objective, as discussed in \ref{d:important}, is to uncover crucial concept evolutions that impact class predictions.
For example, how important is the evolution of a neuron's concept (e.g., from \textit{``furry animals' eyes''} to \textit{``human neck''}) to the prediction for a class (e.g., ``bow tie'')?
Inspired by~\cite{kim2018interpretability}, we quantify the significance of a concept evolution by evaluating how sensitive a class prediction is to the evolutionary state of the concepts.

Eq~\eqref{eq:sensitivity} defines such sensitivity of the class $c$ prediction with respect to the concept evolution of neuron $n$ in layer $l$ in model $M$, from epoch $t$ to $t'$, given an input $\mathbf{x}$.
$Z_{l,M}^{t}(\mathbf{x})$ is the activation map of all neurons in $l$ at $t$ for $\mathbf{x}$.
The function $h^t_{l,M,c}(\cdot): \mathbb{R}^{h_l \times w_l \times s_l} \rightarrow \mathbb{R}$ takes $Z_{l,M}^{t}(\mathbf{x})$ as input and provides the logit value for class~$c$, where $h_l$, $w_l$, and $s_l$ are height, width, and the number of neurons in $l$, respectively.
$\Delta Z_{n,l,M}^{t, t'}(\mathbf{x})$ is the activation change of $n$ from $t$ to $t'$, as defined in Eq~\eqref{eq:actchange}, where $\mathbf{0}_{a,b}$ is a zero matrix of $a$ rows and $b$ columns.
The directional derivative in Eq~\eqref{eq:sensitivity} indicates how sensitively a prediction for class $c$ would change if the activation in layer $l$ changes towards the direction of neuron $n$'s evolution.
A positive value indicates that the concept evolution of neuron $n$ positively contributes to the prediction for class $c$.

\begin{equation}
    \label{eq:actchange}
    \Delta Z_{n,l,M}^{t, t'}(\mathbf{x})
        = [ 
            \mathbf{0}_{h_l,w_l}, 
            \cdots , 
            \underbrace{
                Z_{n,l,M}^{t'}(\mathbf{x}) - Z_{n,l,M}^{t}(\mathbf{x})
            }_{n\text{-th matrix}}, 
            \cdots , 
            \mathbf{0}_{h_l,w_l}
        ]
\end{equation}
\begin{equation}
    \label{eq:sensitivity}
    \begin{split}
    &S_{n, l, M, c}^{t,t'} (\mathbf{x})\\
        &= 
        \lim_{\epsilon \rightarrow 0}
        \frac{
            h^t_{l,M,c}\big(
                Z^{t}_{l,M}(\mathbf{x})
                + \epsilon \Delta Z_{n,l,M}^{t,t'}(\mathbf{x})
            \big)
            - h^t_{l,M,c}\big(Z^{t}_{l,M}(\mathbf{x})\big)
        }{
            \epsilon
        }
        \\
        &= 
        \nabla h^{t}_{l,M,c}(Z^{t}_{l,M}(\mathbf{x})) 
        \cdot 
        \Delta Z_{n,l,M}^{t, t'}(\mathbf{x})
    \end{split}
\end{equation}

We finally measure the importance of concept evolution of a neuron $n$ in layer $l$ in model $M$ from epoch $t$ to $t'$ for class $c$, by aggregating the importance across class $c$ images, as in Eq~\eqref{eq:importance}, where $X_c$ is the set of images labeled as $c$.
\begin{equation}
    \label{eq:importance}
    I^{t, t'}_{n, l, M, c} = 
        \frac{|\{\mathbf{x} \in X_c\;:\; S^{t, t'}_{n, l, M, c}(\mathbf{x}) > 0\}|}{|X_c|}
\end{equation}

\begin{figure}[t]
    \centering
    \includegraphics[width=1\columnwidth]{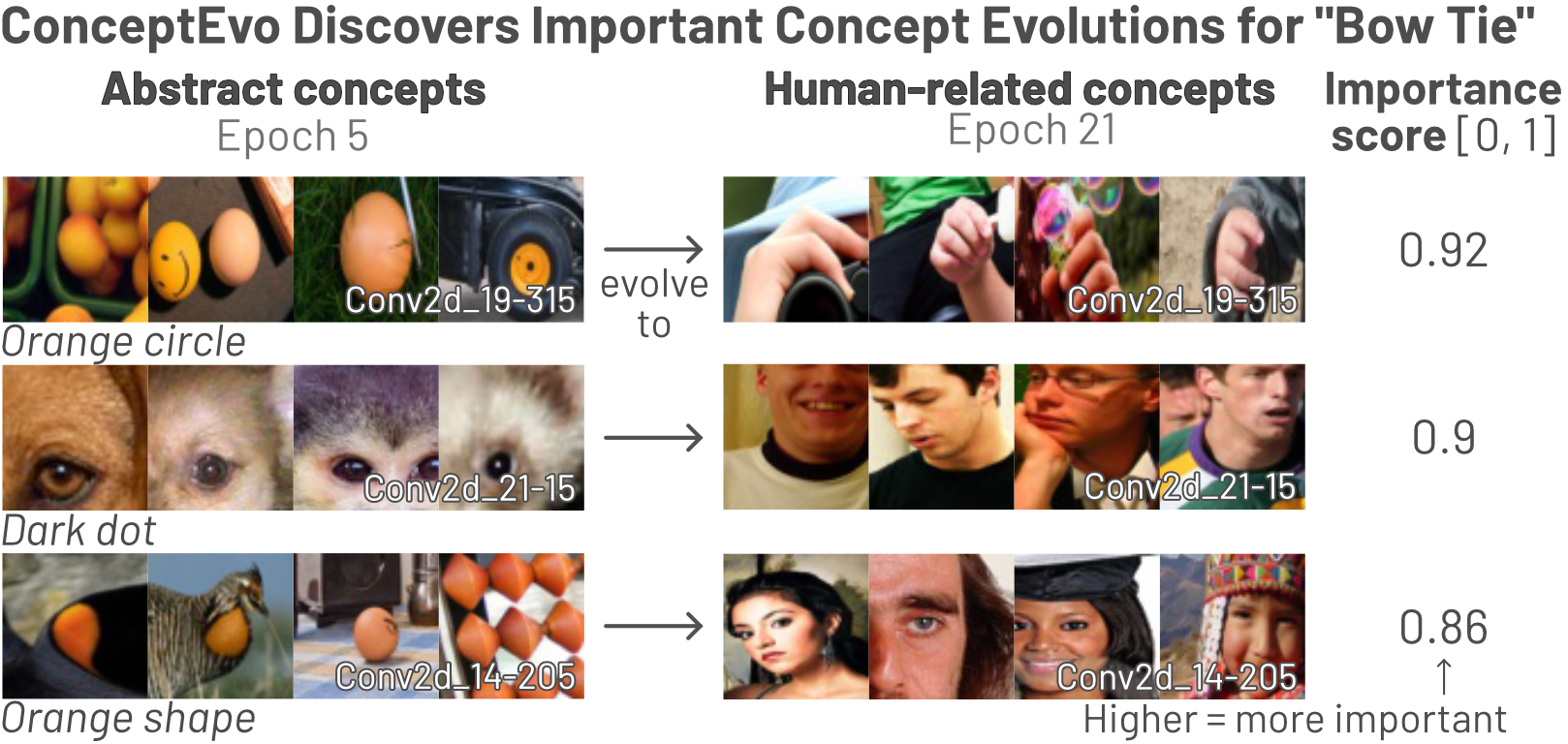}
    \vspace{-2em}
    \caption{
        \model identifies and quantifies important concept evolutions for class prediction.
        For example, in a VGG16, it discovers that concepts evolving towards human-related attributes, such as \textit{``orange circles''} $\rightarrow$ \textit{``hand''} in the top row, are important for the ``bow tie'' class.
        The importance score for this evolution is 0.92, meaning that such a concept evolution enhances predictions for 92\% of bow tie images.
    }
    \vspace{-1em}
    \label{fig:important-evo}
\end{figure}

Fig~\ref{fig:important-evo} illustrates important concept evolutions for the ``bow tie'' class discovered by \model, such as evolutions from abstract concepts to \textit{``hand,''} \textit{``neck,''} and \textit{``face''} concepts.
Surprised by the many evolutions towards human-related concepts, we inspected the raw images for the bow tie class and found that the majority of the images (over 70\%) depict a person wearing a bow tie.

\subsection{Runtime and Time Complexity}

We designed \model with a focus on practicality, considering the need for real-time interpretation during model training.
To ensure this, we aimed to keep the runtime of our approach shorter than a single training epoch, allowing simultaneous training and interpretation.
Our approach meets this requirement. 
Below, we report the runtime of \model when using an NVIDIA A6000 GPU with 40GB RAM and the 10\% randomly sampled ImageNet dataset \cite{russakovsky2015imagenet} with 120 K images.

In the two-step concept evolution interpretation method of \model (described in Sec. \ref{sec:comparable}), 
Step 1, which creates the base semantic space, completes in less than 30 minutes.
Step 2, which unifies the semantic space of models across epochs, takes less than 3 hours for Step 2.1 (image embedding) and less than 1 hour for Step 2.2 (identifying stimuli of a non-base model and approximating the embedding of its neurons).

Step 2.2 ($\sim$1 hour) is the only procedure that needs to be performed when projecting concepts in a new model onto the unified semantic space, and its runtime is shorter than training a model for an epoch (e.g., ConvNeXt takes 1.56 hours). 
This means that \model's interpretation can be performed concurrently with model training.
Step 1 ($\sim$30 minutes) and Step 2.1 ($\sim$3 hours) are one-time computations that can be reused, making \model a practical and efficient choice.

\subsubsection{General Interpretation of Concept Evolution (Sec. \ref{sec:comparable})}

Now, we provide a detailed analysis of the time complexity of \model's two-step concept evolution interpretation method described in Section~\ref{sec:comparable}. 
The ``steps'' mentioned here correspond to the steps outlined in Section \ref{sec:comparable}.

\textit{\textbf{Step 1: Creating the base semantic space.}}
Step 1 has an overall time complexity of $O(|N_{M_b}| \cdot |I|)$, where $N_{M_b}$ is the set of neurons in $M_b$, and $I$ is the set of images. 

In Step 1.1, the time complexity is $O(|N_{M_b}| \cdot |I|)$.
For each neuron, collecting the top $k$ images from $|I|$ images takes $O(|I| \cdot k)$.
This process involves maintaining a sorted list of length $k$, which stores the top-$k$ images observed so far.
At each iteration for an image $\mathbf{x}$, we compare $\mathbf{x}$ to the smallest top-$k$ item in the list.
If $\mathbf{x}$ results in a higher activation for the neuron, we insert $\mathbf{x}$ into the list and remove the previous smallest top-$k$ item.
Identifying the proper spot to insert $\mathbf{x}$ and inserting it (if necessary) takes $O(k)$, and $k$ is small (e.g., 10). 
Thus, the total time for collecting the top $k$ images from $|I|$ images is $O(|I| \cdot k) = O(|I|)$.
Therefore, for all neurons, Step 1.1 has a time complexity of $O(|N_{M_b}| \cdot |I|)$.

In Step 1.2, the time complexity is $O(|N_{M_b}|)$. 
Step 1.2 consists of two sub steps. 
First, for each image $\mathbf{x}$, collecting neurons with $\mathbf{x}$ in their stimuli takes $O(|N_{M_b}|)$, as it requires iterating through all stimuli of all neurons, which is a total of $O(k \cdot |N_{M_b}|)$.
Second, for each image $\mathbf{x}$ and its corresponding co-activated neurons, sampling neuron pairs from the list of co-activated neurons with the sliding window takes $O(k \cdot |N_{M_b}|) = O(|N_{M_b}|)$.
This results in $O(|N_{M_b}|)$ pairs of neurons.
The sampling process is repeated $E$ times, thus the total time for Step 1.2 is $O(E \cdot (|N_{M_b}| + |N_{M_b}|)) = O(|N_{M_b}|)$. 

Step 1.3 takes $O(|N_{M_b}|)$, as the number of generated neuron pairs in Step 1.2 is $O(|N_{M_b}|)$. 
One epoch of gradient descent in Step 1.3 takes $O(|N_{M_b}| \cdot R) = O(|N_{M_b}|)$ , resulting in a final time complexity of $O(|N_{M_b}|)$.

Overall, the time complexity of Step 1 is $O(|N_{M_b}| \cdot |I|)$ + $O(|N_{M_b}|) + O(|N_{M_b}|)$ = $O(|N_{M_b}| \cdot |I|)$. 
One advantage of this approach is its linear time complexity with respect to the number of neurons, instead of quadratic time. 
This is because it avoids the need to compare and represent concepts for all pairs of neurons, and instead focuses on sampled pairs of neurons.

\textit{\textbf{Step 2: Unifying the semantic space.}}
Overall, Step 2 has a time complexity of $O(|N_{M_b}| \cdot |I|)$.
In Step 2.1, the time complexity is $O(|N_{M_b}| \cdot |I|)$.
This is because optimizing $J_2$ takes $O(|I|)$ time to learn $O(|I|)$ vectors, and approximately representing images not covered by the base model's stimuli also takes $O(|N_{M_b}| \cdot |I|)$, similar to Step 1 (since it adapts Step 1).
To represent the concepts of neurons in a non-base model $M$ within the unified semantic space, Step 2.2 takes $O(|N_{M}| \cdot |I|)$.
This step involves computing stimuli for each neuron in $M$, where $N_M$ is the neurons in $M$, using a similar approach as in Step 1.1.

\subsubsection{Concept Evolution Important for a Class (Sec. \ref{sec:important})}
Finding important concept evolutions for each class $c$ requires $O(|I| \cdot |N_{M_b}|)$ time, since the computation of neuron sensitivity (Eq \eqref{eq:sensitivity}) relies on the number of images labeled as $c$ (which is $O(|I|)$).
In terms of runtime, on average, this process took 37 minutes for VGG16, InceptionV3, and ConvNeXt models.

\section{Experiment}
\label{sec:experiment}

We evaluate how well \model satisfies the desired properties for interpreting concept evolution (Sec \ref{sec:desiderata}, D1-3) by addressing the following research questions:

\begin{enumerate}[label=\textbf{Q\arabic*},itemsep=0mm, topsep=0mm,parsep=0mm, leftmargin=6mm]
    \item \textbf{Alignment.} 
        How effectively does \model align concepts of different models at different training stages in the unified semantic space? 
        (Sec \ref{sec:alignability}, for \ref{d:comparable})
    \item \textbf{Meaningfulness.} 
        To what extent are the discovered concept evolutions semantically meaningful?
        (Sec~\ref{sec:meaningfulness}, for \ref{d:comparable})
    \item \textbf{Importance.} 
        How important are the discovered concept evolutions in terms of their impact on class prediction?
        (Sec \ref{sec:eval-important}, for \ref{d:important})
    \item \textbf{Discoveries.} 
        How does \model contribute to the discovery of insightful findings?
        (Sec \ref{sec:discovery}, for \ref{d:actionable})
\end{enumerate}
\subsection{Experiment Settings}
\label{sec:exp-setting}

\textbf{Datasets and models.} 
We examine concept evolutions in representative image classifiers trained on ILSVRC2012 (ImageNet)~\cite{russakovsky2015imagenet}.
The models we investigate include a modern model, such as ConvNeXt \cite{liu2022convnet} which draws inspiration from recent architectures such as ResNet \cite{targ2016resnet}, ResNeXt~\cite{xie2017aggregated}, and vision transformers \cite{liu2021swin, vit2021}.
Additionally, we investigate classic models such as VGG16~\cite{simonyan2015very}, VGG19~\cite{simonyan2015very}, VGG16 without dropout layers~\cite{srivastava2014dropout}, and InceptionV3 \cite{szegedy2016rethinking}.
To ensure comparable accuracies, we trained these models using the hyperparameters reported in prior work \cite{simonyan2015very, szegedy2016rethinking, liu2022convnet}.

\textbf{Hyperparameter settings.} 
We selected hyperparameters to achieve the overarching goal of a unified semantic space that balances strong coherence among neighboring neurons with computation efficiency.
Specifically, the following hyperparameters were tested within the indicated ranges:
the number of stimuli per neuron ($k$) 
was tested from 5 to 30, with a chosen value of 10 to strike the balance;
the dimension of neuron and image embeddings was set to 30 (tested from 5 to 100);
the learning rate for neuron embedding was set to 0.05 and for image embedding, it was set to 0.1 (tested from 0.001 to 0.5);
and the number of randomly sampled neurons per neuron pair ($R$) was set to 3 (tested from 0 to 5).
\vspace{-1em}
\subsection{Alignment of Neuron Embeddings}

\label{sec:alignability}
To ensure the effectiveness of \model in aligning concepts across models and epochs, we conducted a large-scale human evaluation using Amazon Mechanical Turk (MTurk), following the methodology of prior work \cite{park2021neurocartography, ghorbani2019towards}. 
The evaluation focused on four categories:
(1) hand-picked sets of neurons representing similar concepts, which served as a baseline;
(2) neuron groups detected by \model from the \basemodel (a well-trained VGG16); 
(3) neuron groups in the same model at different training epochs, detected by \model;
(4) neuron groups from different models at different epochs, detected by \model.
To collect the neuron groups, we applied K-means clustering on the neuron embeddings within the unified semantic space.

\begin{figure}[t]
    \centering
    \includegraphics[width=1\columnwidth]{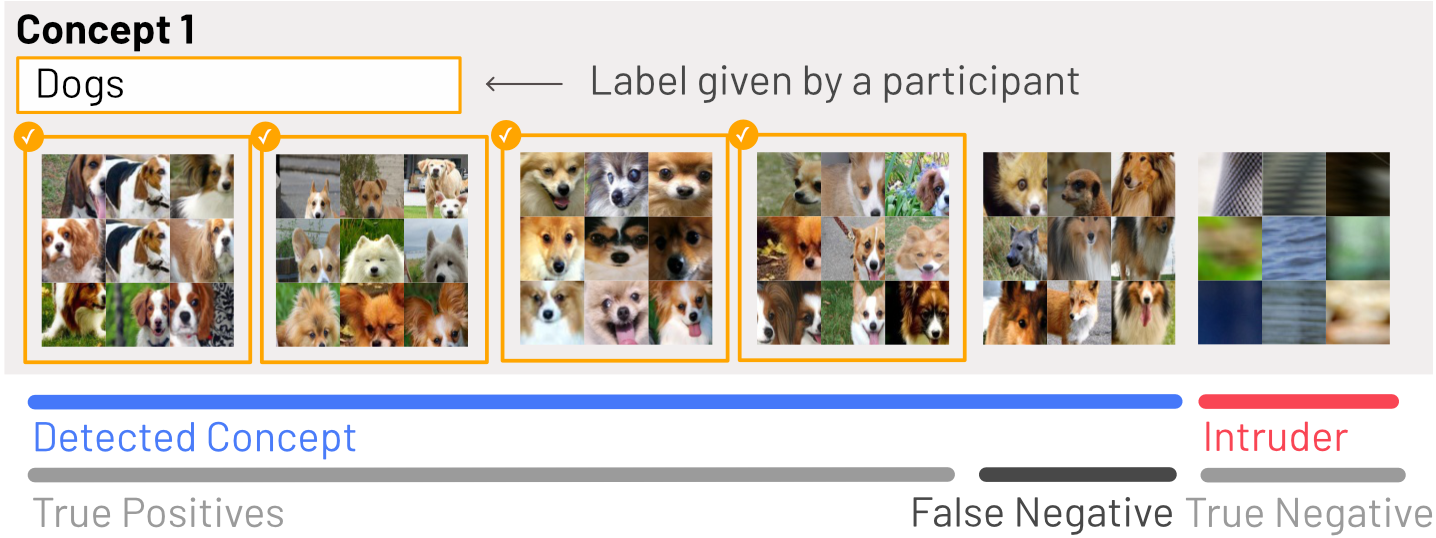}
    \vspace{-1em}
    \caption{
        MTurk questionnaire example. 
        Participants are presented with six neurons' example patches and asked to determine if they are a semantically coherent group.
        If they identify a coherent group, they provide a short label for that group.
        In the provided example, the first five neurons are semantically similar, detected and grouped by \model.
        The rightmost is randomly sampled and unrelated to others.
        Here, a participant correctly identifies the first four neurons as a coherent ``dogs'' concept (four \textit{true positives}), misses the fifth neuron (one \textit{false negative}), and correctly identifies the intruder as unrelated (one \textit{true negative}).
    }
    \label{fig:survey}
\end{figure}

\begin{figure}[t]
    \centering
    \includegraphics[width=0.8\columnwidth]{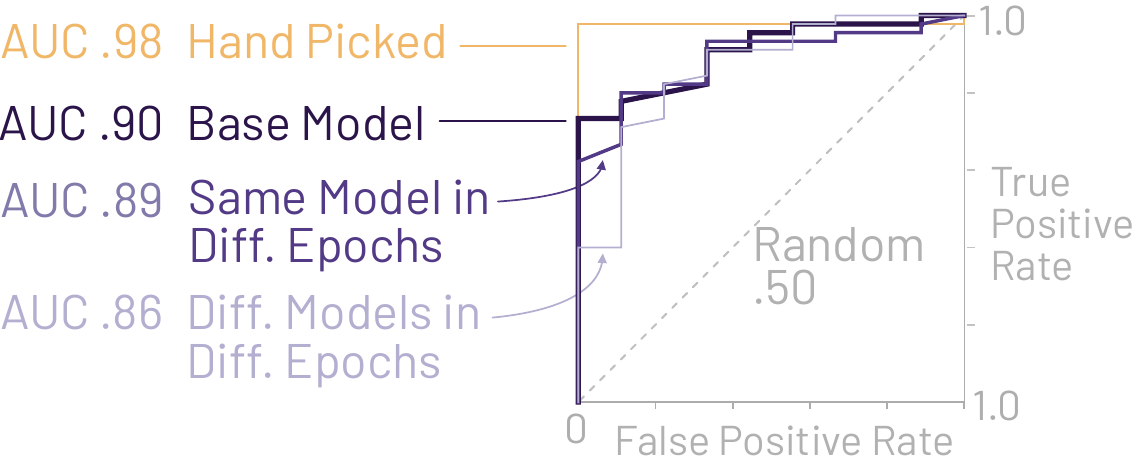}
    \vspace{-1em}
    \caption{
        ROC Curve for human estimations demonstrating the high alignability of concepts discovered by \model, even when sampled across different models and epochs.
    }
    \label{fig:roc}
\end{figure}

We conducted concept classification tasks with 260 MTurk participants, where each participant completed nine unique tasks.
Each task consisted of six neurons presented in random order, where five of them had similar concepts identified by \model or were hand-picked, while one neuron served as a randomly selected ``intruder'' neuron. 
To help participants understand the concept of each neuron, we provided nine example image patches.
Participants were not informed about the potential presence of intruders and were asked to select as many neurons as they believed to be semantically similar.
They were also asked to provide a brief description of the concept they perceived. 
This process, as illustrated in Fig~\ref{fig:survey}, essentially forms a classification task, treating the participants as classifiers and the grouped neurons as true labels.
A total of 10,950 individual classification tasks were generated for the test set.
From this framing, we consider success based on the level of agreement of participants with the model's determination.
Fig \ref{fig:roc} shows an ROC curve with the participants' determinations, demonstrating the high discernibility and alignment of \model-detected concepts.
Even when sampling concepts across different epochs and models, the AUC scores remain consistently high, ranging from 0.90 for sampling within the base model to 0.86 for sampling across different models and training epochs.

\subsection{Meaningfulness of Concept Evolution}
\label{sec:meaningfulness}
Concepts discovered by \model should be meaningful and informative to humans.
We evaluate the \textit{interpretive consistency} of the concepts labeled and described by the participants, as shown in Fig \ref{fig:survey}.
To handle variations in phrasing for the labels, we use sentence-level embeddings from the Universal Sentence Encoder (USE)~\cite{cer2018universal}. 
USE captures the semantic similarity between phrases, such as ``vehicle wheels,'' ``cars,'' and ``trucks'', which should have high USE similarity.
To establish a baseline for similarity, we calculate the average pairwise similarity between all labels, resulting in a value of 0.28.
Subsequently, we measure the average pairwise similarity between the labels provided by participants for individual concepts within each category from \ref{sec:alignability}.
The results are as follows:
(1) the average concept similarity for hand-picked concepts is 0.455,
(2) the average concept similarity for concepts from the base model is 0.40,
(3) the average concept similarity for concepts within the same model but different epoch is 0.40, and
(4) the average concept similarity for concepts from different models and different epochs is 0.38.
All of these values significantly exceed the baseline similarity value of 0.28. 
This indicates that the concepts discovered through \model are reliable and meaningful, even when assessed by different people.

\begin{figure}[t]
    \centering
    \includegraphics[width=1\columnwidth]{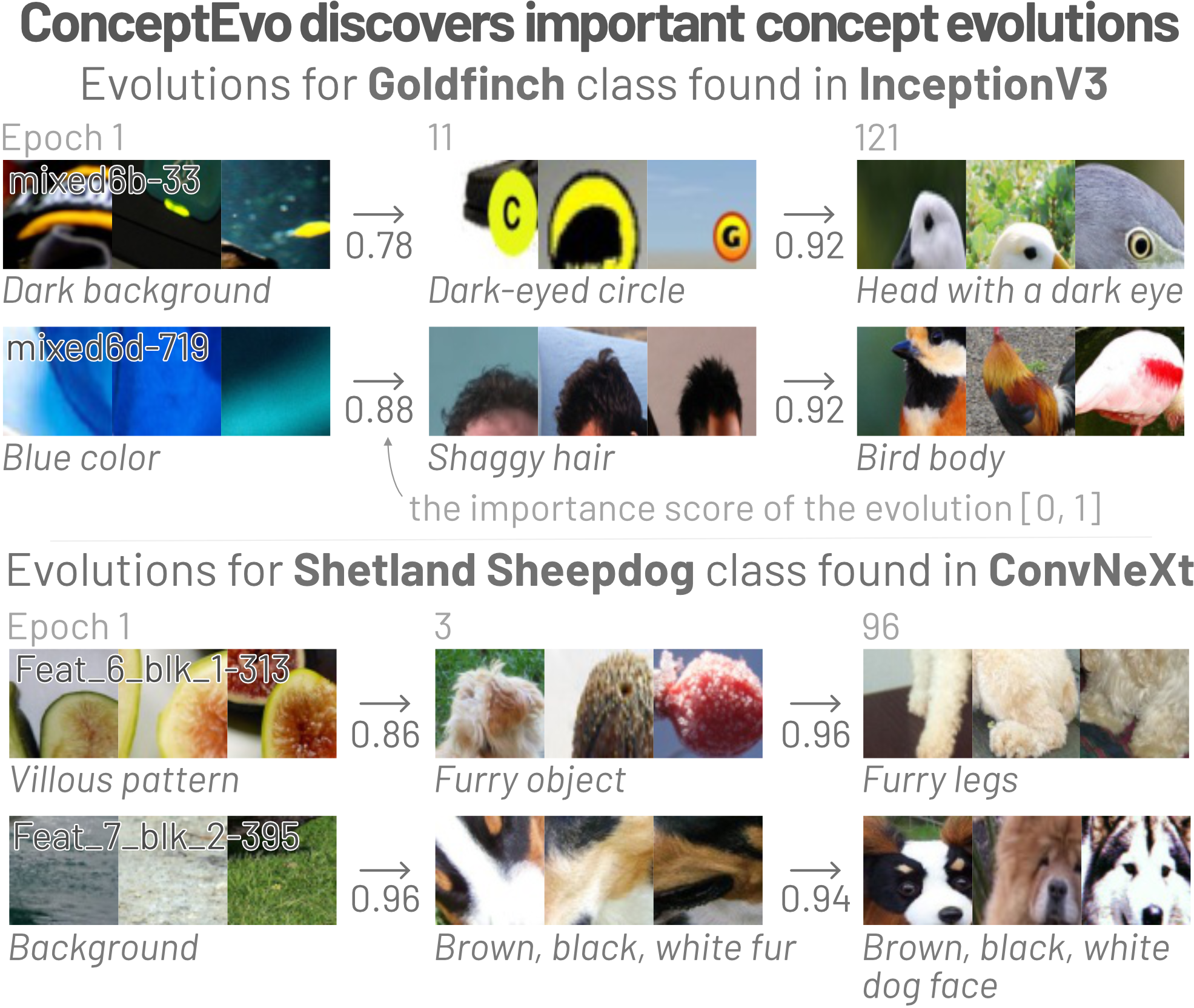}
    \vspace{-2em}
    \caption{
        \model discovers concept evolutions important for class predictions.
        For example, it discovers 
        bird-related evolutions important for the ``Goldfinch'' class in InceptionV3, and
        dog-related evolutions important for the ``Shetland sheepdog'' class in ConvNeXt.
        Some neurons become increasingly specialized as training progresses.   
        For example, in the first row, a neuron that initially detects abstract concept of \textit{dark background} evolves to detect \textit{dark-eyed circle}, and then further evolves to detect \textit{head with a dark eye}.
    }
    \vspace{-1em}
    \label{fig:imp-evo-example}
\end{figure}
\subsection{Concept Evolutions Important to a Class}
\label{sec:eval-important}

\model quantifies and identifies important concept evolutions, as illustrated in Fig \ref{fig:imp-evo-example}.
In InceptionV3, it reveals evolutions from abstract concepts to bird-related concepts that aid in classifying the ``Goldfinch'' class.
Similarly, in ConvNeXt, it discovers evolutions from abstract concepts to dog-related concepts that are important for classifying the ``Shetland sheepdog'' class. 
As training progresses, some neurons become more specialized.
For example, in the first row of Fig~\ref{fig:imp-evo-example}, 
a neuron initially detecting abstract concepts of a \textit{dark background} evolves to detect a \textit{dark-eyed circle} and later to detect a \textit{head with a dark eye}.

To evaluate the effectiveness of \model in discovering important concept evolutions, we measure the changes in accuracy when evolutions are reverted, similar to how prior work evaluated concept importance in fully-trained models \cite{ghorbani2020neuron, ghorbani2019towards}. 
By reverting a neuron's activation map from $t'$ to $t$, we evaluate the prediction accuracy at $t'$. 
A larger drop in accuracy indicates a higher importance for the concept evolution of that neuron.
To determine the stages of evolution to evaluate, we identify the epochs with the closest top-1 training accuracies to the milestones of 25\%, 50\%, and 75\%.
Specifically, for VGG16, the evolution stages are 5$\rightarrow$21 and 21$\rightarrow$207;
for InceptionV3, 1$\rightarrow$11 and 11$\rightarrow$121; and
for ConvNeXt, 1$\rightarrow$3 and 3$\rightarrow$96.

\begin{figure}[t]
    \centering
    \includegraphics[width=0.9\columnwidth]{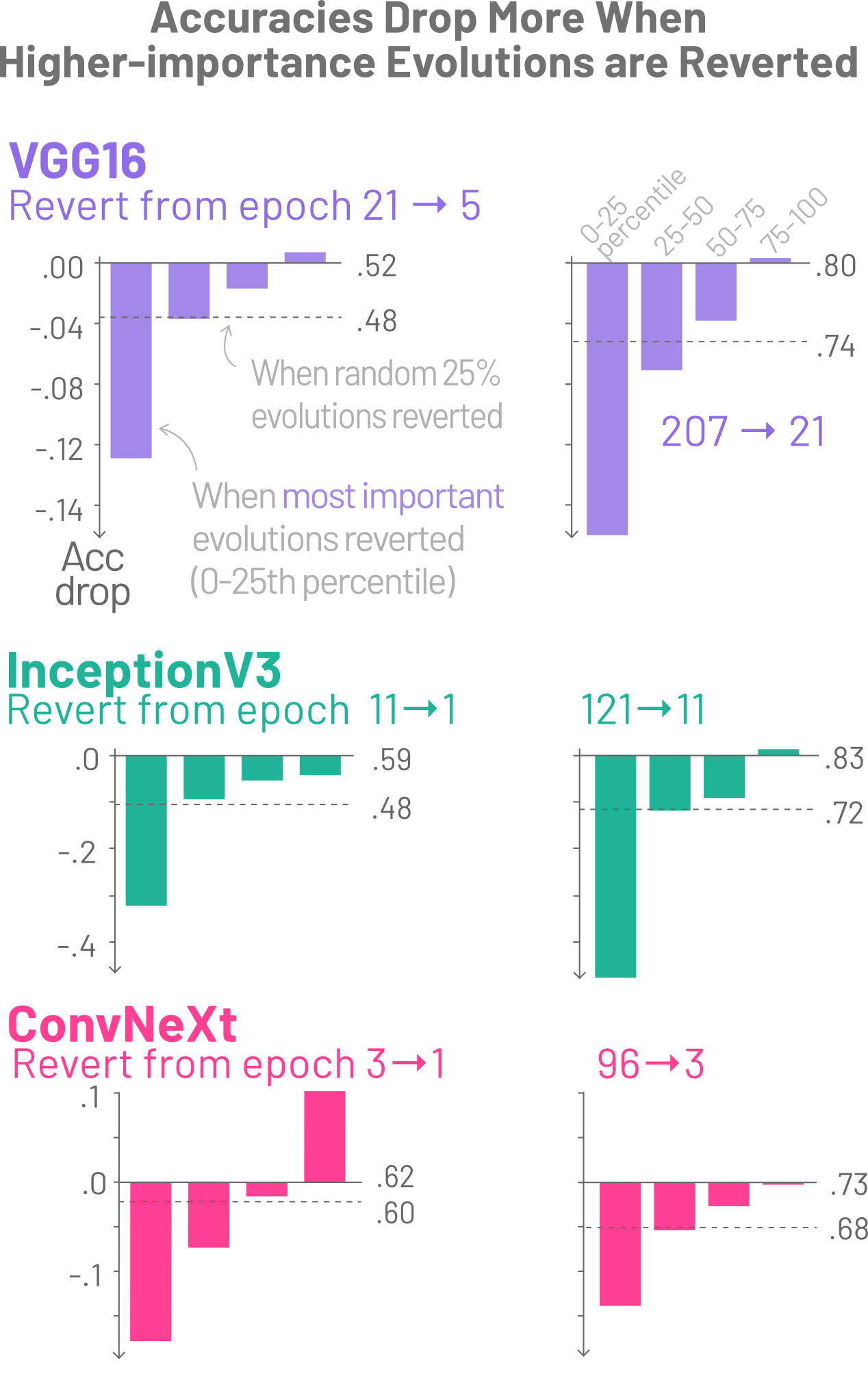}
    \vspace{-1em}
    \caption{
        We evaluate the ability of \model to quantify and identify important concept evolutions for 100 randomly selected classes.
        Neurons are ranked based on their evolution importance and then divided into four bins: 0-25th (most important), 25-50th, 50-75th, 75-100th percentiles.
        By reverting higher-importance evolutions, we observed a larger drop in top-1 training accuracy, demonstrating the effectiveness of \model in quantifying and identifying important concept evolutions.
        As a baseline, for comparison, we also measured the accuracy drop when randomly reverting 25\% (i.e., the same number of neurons in each bin) evolutions, which fell between the 25-50th and 50-75th percentile bins.
    }
    \vspace{-1em}
    \label{fig:eval-imp-train1}
\end{figure}

As \model measures the importance of concept evolution for a single neuron (as defined in Eq \ref{eq:importance}), it is natural to evaluate accuracy changes by reverting each neuron's evolution individually and then aggregating the changes.
However, due to the large number of neurons, this approach becomes computationally prohibitive.
To address this, we propose a more practical approach that reverts multiple evolutions in a layer at a time and aggregates the accuracy changes across layers.
The evaluation process consists of five steps for each class $c$ and evolution stage from epoch $t$ to $t'$.
\textbf{Step~1:} Sample 128 images for class $c$, which corresponds to approximately 10\% of the total images for that class (around 1300 images).
\textbf{Step~2:} Compute the importance of concept evolutions for all neurons, using Eq~\eqref{eq:importance}.
\textbf{Step 3:} Rank the neurons in each layer based on their evolution importance and divide them into four importance bins: 0-25th percentile (most important), 25-50th percentile, 50-75th percentile, and 75-100th percentile.
\textbf{Step 4:} Revert the evolutions of neurons in each bin, compute the accuracy at epoch $t'$, and measure the accuracy changes compared to the non-reverted accuracy.
\textbf{Step 5:} Average the accuracy changes across layers to obtain the accuracy changes for the four bins.
To mitigate sampling bias in Step 1, we repeat the above procedure five times independently.
We average the accuracy changes across 100 randomly selected classes from the 1,000 classes in ImageNet\footnote{Standard deviations of the average accuracy changes across the classes between the five runs are very low (e.g., 9.2e-5 for top-1 training accuracy and 2.1e-4 for top-1 test accuracy, for the 21$\rightarrow$207 evolution).}.

Fig~\ref{fig:eval-imp-train1} illustrates the impact of reverting evolutions in different importance bins on the top-1 training accuracy of VGG16, InceptionV3, and ConvNeXt.
Notably, reverting higher-importance evolutions (lower percentiles) results in larger accuracy drops, confirming the effectiveness of \model in quantifying and identifying important concept evolutions.
Interestingly, reverting the least important evolutions (75-100th percentile) sometimes leads to increased accuracy. 
This suggests that the least important evolutions may interfere with the corresponding class predictions.
As a baseline, we reverted 25\% randomly selected evolutions, resulting in an accuracy drop between the 25-50th percentile and the 50-75th percentile.
Furthermore, we evaluated the changes in the top-5 training, top-1 test, and top-5 test accuracies when reverting evolutions in the same four bins, reinforcing our key finding that reverting higher-importance evolutions results in a larger accuracy drop.

\subsection{Discovery}
\label{sec:discovery}

\begin{figure}[t!]
    \centering
    \includegraphics[width=0.65\columnwidth]{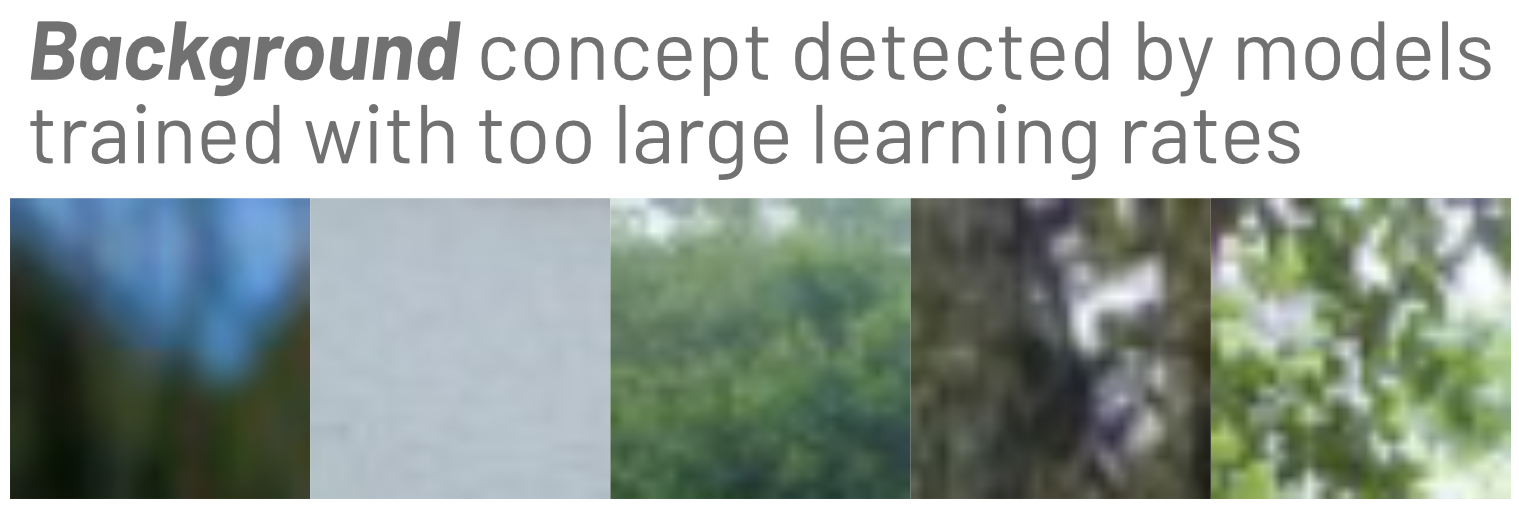}
    \caption{
        An example of \textit{``background''} concept detected by VGG16 and ConvNeXt that are trained with overly large learning rates, when the accuracy is very low.
        In the last convolutional layer in these models, a notable percentage (over 30\%) of neurons show exclusive intense activation in response to backgrounds of images.
    }
    \label{fig:bg-concept}
\end{figure}

\begin{figure}[t!]
    \centering
    \includegraphics[width=0.95\columnwidth]{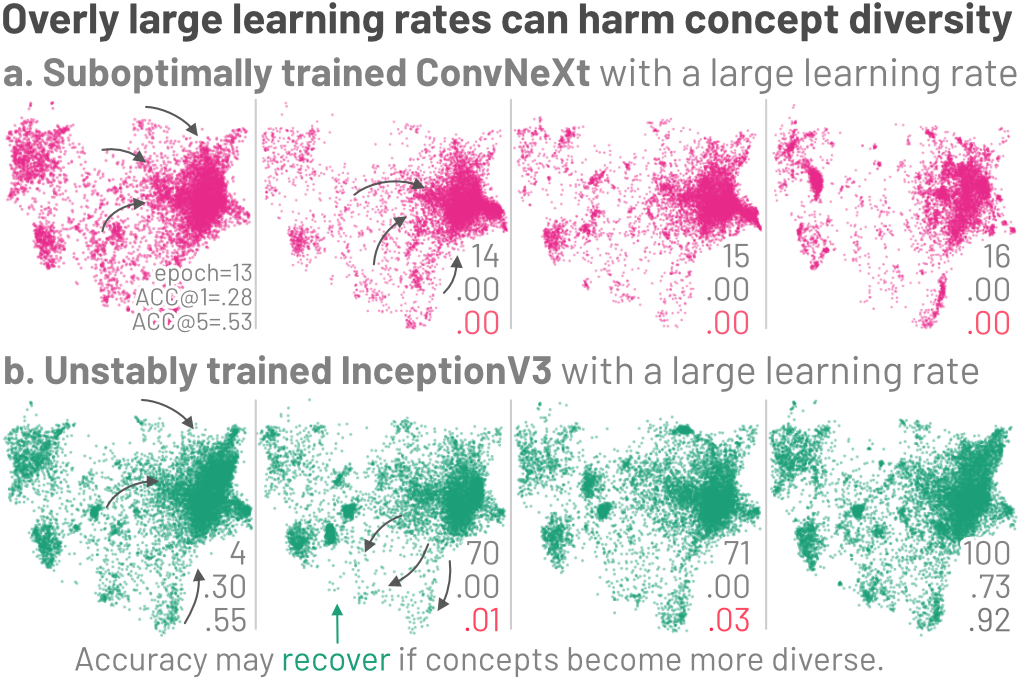}
    \caption{
        A suboptimally trained ConvNeXt and an unstably trained InceptionV3 with large learning rate experience decreased concept diversity and convergence in certain regions (e.g., right side to detect lower-level concepts), specifically when these models' training accuracies drop (as seen in the second column).
        Interestingly, the training accuracy of InceptionV3 recovers, because the concepts become more diverse starting from epoch 70, showing a better recovery resilience.
    }
    \vspace{-1em}
    \label{fig:discovery-large-learning-rate}
\end{figure}

\textbf{Incompatible hyperparameters harm concept diversity.}
\model's aligned neuron concept embedding
helps identify problems caused by incompatible hyperparameters and offer insights into their impact on model performance.
For example, in Fig~\ref{fig:discovery}b, \model reveals that a VGG16 suboptimally trained with an excessively high learning rate\footnote{0.05, larger than an optimal learning rate 0.01 presented in prior work} exhibits a drastic accuracy drop over training epochs. 
Early signs of problems, such as the ``atrophying'' of neuron concepts that degrade concept diversity and only detect lower-level concepts, become apparent even before the accuracy reaches~0.
The loss of diversity is so severe that it cannot be recovered even with 40 additional training epochs.
A similar pattern is observed in a ConvNeXt model trained with a high learning rate\footnote{0.02, larger than an optimal learning rate 0.004 used in prior work}, as shown in Fig~\ref{fig:discovery-large-learning-rate}a.
In cases where the accuracy is low in VGG16 and ConvNeXt, we observe a significant reduction in concept diversity, especially in the last convolutional layers.
For example, as seen in Fig~\ref{fig:bg-concept}, almost all neurons in VGG16 and over 30\% of neurons in ConvNeXt predominantly detect \textit{``background''} concepts.

In the case of an InceptionV3 unstably trained with a large learning rate\footnote{1.5, larger than an optimal rate of 0.045 used in prior work}, \model reveals a similar yet slightly different scenario. 
As depicted in Fig~\ref{fig:discovery-large-learning-rate}b, the accuracy significantly drops at epoch 70, but interestingly, it recovers after a few more epochs.
This recovery is likely due to the persistence of a large number of concepts at epoch 70 and the increasing diversity of concepts, despite the low accuracy.

These examples demonstrate that \model can provide actionable insights to determine whether interventions, such as stopping the training, might be beneficial.
Severe damage to concept diversity, as observed in Fig \ref{fig:discovery}b and \ref{fig:discovery-large-learning-rate}a, suggests that stopping the training might be more beneficial, as the model is unlikely to recover even with further epochs, compared to 
a better ability to recover the concept diversity as depicted in Fig~\ref{fig:discovery-large-learning-rate}b.

To quantitatively study concept diversity, we use differential entropy which measures the uncertainty in a continuous variable \cite{michalowicz2013handbook}.
We compute the differential entropy for each dimension of neuron embeddings and average the values across the dimensions\footnote{We average the differential entropy across reduced 2D embeddings, instead of the original dimension, since computing the differential entropy for some high dimensional vectors leads to infinity.}. 
Higher values indicate more diverse concepts.
In a VGG16 suboptimally trained with a large learning rate (Fig \ref{fig:discovery}b), the differential entropy decreases: 
1.89$\rightarrow$1.48$\rightarrow$-1.80$\rightarrow$-1.80 for epochs 3, 12, 13, 14,
indicating a loss of concept diversity.
Similarly, in a suboptimally trained ConvNeXt (Fig~\ref{fig:discovery-large-learning-rate}a), the differential entropy decreases:
1.83$\rightarrow$1.64$\rightarrow$1.59$\rightarrow$1.52 for epochs 13, 14, 15, 16.
In contrast, optimally trained models show increasing differential entropy, indicating that concepts become more diverse over epochs.
For example, in an optimally trained VGG16 (Fig \ref{fig:discovery}a), the differential entropy increases: 1.10$\rightarrow$1.90$\rightarrow$2.06$\rightarrow$2.09 for epochs 0, 5, 21, 207.
In the case of an unstably trained InceptionV3 (Fig \ref{fig:discovery-large-learning-rate}b), the differential entropy decreases until epoch 70 (lowest accuracy) and then rebounds:
1.82$\rightarrow$1.32$\rightarrow$1.54$\rightarrow$1.80 for epochs 4, 70, 71, 100, indicating that its concept diversity was initially damaged but later restored.

\textbf{Overfitting slows concept evolution.}
Overfitting is a common issue in DNN training \cite{rice2020overfitting,cogswell2015reducing}.
Using \model, we have discovered that concepts in overfitted models evolve at a slower pace, despite experiencing rapid increases in training accuracy.
To intentionally induce overfitting, we modified a VGG16 (Fig~\ref{fig:discovery}c) by removing its dropout layers which are known to help mitigate overfitting \cite{srivastava2014dropout}.
Additionally, we overfit a ConvNeXt model by setting the weight decay of the AdamW optimizer to 0, reducing its regularization effect \cite{loshchilov2017decoupled}.
These models are overfitted expectedly\footnote{In VGG16, at epoch 30, its top-1 train, top-5 train, top-1 test, top-5 test accuracies are 0.99, 1, 0.37, 0.61, respectively.
In ConvNeXt, at epoch 32, its top-1 train, top-5 train, top-1 test, top-5 test accuracies are 0.94, 0.99, 0.57, 0.80, respectively.}. 

We observed that overfitted models show slower concept evolution compared to their corresponding well-trained models.
To increase the top-1 training accuracy from approximately 0.25 to 0.5 and from approximately 0.5 to 0.75, the neuron embeddings in a well-trained VGG16 model (Fig \ref{fig:discovery}a) move an average Euclidean distance of 2.08e-4 and 2.90e-4, respectively.
In contrast, the overfitted VGG16 model (Fig \ref{fig:discovery}b) exhibits much slower movement, with neuron embeddings only shifting by 1.94e-4 and 1.76e-4 for the same accuracy increments.
Similarly, for the well-trained ConvNeXt model, raising the top-1 training accuracy from approximately 0.25 to 0.5 and from approximately 0.5 to 0.75 corresponds to neuron embeddings moving an average distance of 1.49e-4 and 1.33e-4, respectively. Conversely, the overfitted ConvNeXt model shows slower movement, with neuron embeddings shifting by only 1.48e-4 and 1.27e-4 for the same accuracy increments.

\vspace{-1em}
\subsection{Comparison with Existing Approaches}

We compare \model with existing methods for representing evolving concepts.
Existing methods are not optimized to capture changes across epochs; 
they can only be applied to one epoch at a time, independently of other epochs.
In our comparison, we consider NeuroCartography~\cite{park2021neurocartography} 
and ACE~\cite{ghorbani2019towards}. 
ACE represents concepts using image segments that activate a layer.
We use the final layer to follow the approach described in the original work.
For image segments, we use the Broden dataset~\cite{bau2017network}.
For 2D visualization of concepts, we use UMAP~\cite{mcinnes2018umap}.
To ensure alignment across epochs, we run UMAP for all epochs simultaneously, avoiding misalignment caused by independent epoch-based reduction.

\begin{figure}[t]
    \centering
    \includegraphics[width=0.9\columnwidth]{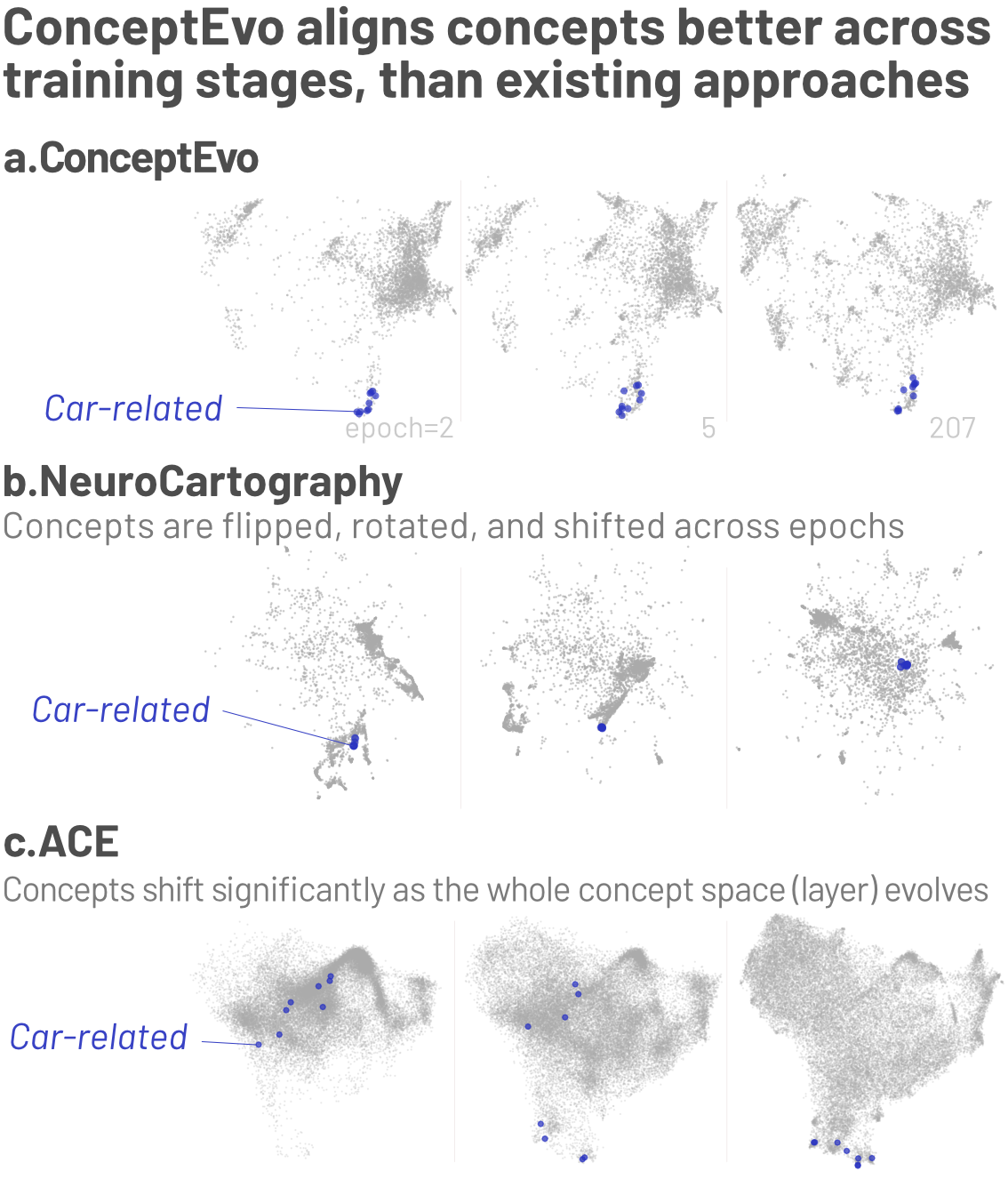}
    \vspace{-1em}
    \caption{
        We compare the representation of concepts in VGG16 using ConceptEvo with existing methods.
        (a) The results show that \model effectively aligns learned concepts across training epochs, by projecting similar concepts to similar embedding locations.
        (b) In contrast, concepts represented by NeuroCartography exhibit flipping, rotation, and shifting  across epochs, indicating misalignment.
        (c) Similarly, concepts represented by ACE undergo significant shifting, as the entire concept space (layer activation space) changes during training, indicating misalignment as well.
    }
    \vspace{-2em}
    \label{fig:baseline-alignment}
\end{figure}

The results show that \model effectively aligns concepts across epochs, while existing methods exhibit misalignment.
In Fig \ref{fig:baseline-alignment}a, the ``car-related'' concept neurons consistently appear at the bottom in epochs 2, 5, and 207. 
In contrast, Fig \ref{fig:baseline-alignment}b demonstrates that the ``car-related'' neurons represented by NeuroCartography exhibit flipping, rotation, and shifting across epochs. 
Similarly, Fig \ref{fig:baseline-alignment}c shows that the ``car-related'' image segments represented by ACE exhibit significant shifting as the concept space changes during training.

\section{Conclusion and Future Work}

\model is a unified interpretation framework for DNNs that reveals the inception and evolution of detected concepts during training.
Through both large-scale human experiments and quantitative analyses, we have showcased the effectiveness of \model in discovering concept evolutions that facilitate human interpretation of model training across different models. 
This framework not only aids in identifying potential training problems but also provides guidance for interventions to achieve more stable and effective training outcomes.

In our future work, we plan to expand the scope of our investigation to include other types of models, such as object detectors, reinforcement learning systems, and language models. 
Additionally, we aim to enhance the alignment of concepts across different models during training. 
Currently, our framework operates under the assumption that an image can be represented by linear combinations of various neurons. 
However, more complex relationships may exist beyond linear associations. 
Thus, we aspire to improve the concept alignment by considering these non-linear relationships, enabling a more comprehensive and accurate representation of concepts across different models.

\section{Acknowledgments}
This work was supported in part by Cisco, DARPA GARD, J.P. Morgan PhD Fellowship, 
NSF \#2144194, gifts from Amazon, Avast, Fiddler Labs, Bosch, Facebook, Intel, NVIDIA, Google, Symantec.

\bibliographystyle{ACM-Reference-Format}
\bibliography{ref}

\end{document}